\documentclass[final,3p,times,twocolumn]{elsarticle}
\usepackage{soul}
\usepackage{xcolor}

\usepackage{amssymb}
\usepackage{url}
\usepackage{amsmath}
\usepackage{todonotes}

\usepackage{booktabs} % For \toprule, \midrule, \bottomrule
\usepackage{multirow} % For multi-column and multi-row cells

\begin{document}

\begin{frontmatter}



\title{Deep Learning tools to support deforestation monitoring in the Ivory Coast using SAR and Optical satellite imagery}


\author[unito]{Gabriele Sartor} %% Author name
\author[unito]{Matteo Salis} %% Author name
\author[unito]{Stefano Pinardi} %% Author name
\author[unito]{Özgür Saracik} %% Author name
\author[unito]{Rosa Meo} %% Author name

\affiliation[unito]{organization={Università degli Studi di Torino},%Department and Organization
            addressline={Computer Science Department}, 
            city={Torino},
            postcode={10149}, 
            state={Italy},
            country={}}

%% Abstract
\begin{abstract}
Deforestation is gaining an increasingly importance due to its strong influence on the sorrounding environment, especially in developing countries where population has a disadvantaged economic condition and agriculture is the main source of income.
In Ivory Coast, for instance, where the cocoa production is the most remunerative activity, it is not rare to assist to the replacement of portion of ancient forests with new cocoa plantations.
In order to monitor this type of deleterious activities, satellites can be employed to recognize the disappearance of the forest to prevent it from expand its area of interest.
In this study, Forest-Non-Forest map (FNF) has been used as ground truth for models based on Sentinel images input.
State-of-the-art models U-Net, Attention U-Net, Segnet and FCN32 are compared over different years combining Sentinel-1, Sentinel-2 and cloud probability to create forest/non-forest segmentation.
Although Ivory Coast lacks of forest coverage datasets and is partially covered by Sentinel images, it is demonstrated the feasibility to create models classifying forest and non-forests pixels over the area using open datasets to predict where deforestation could have occurred.
Although a significant portion of the deforestation research is carried out on visible bands, SAR acquisitions are employed to overcome the limits of RGB images over areas often covered by clouds.
Finally, the most promising model is employed to estimate the hectares of forest has been cut between 2019 and 2020.
\end{abstract}


% \begin{highlights}
% \item Deforestation is a complex phenomenon to be monitored, given the scarcity of global datasets. In addition, Ivory Coast is poorly covered by free satellite imagery.
% \item We propose a framework to detect deforestation using FNF and Sentinel data.
% \item Using Sentinel as input, the output of the framework can be an FNF with a sub-annual rivisit time.
% \end{highlights}

%% Keywords

\end{frontmatter}



\section{Introduction}
\label{sec1}
%% Labels are used to cross-reference an item using \ref command.

%%% Deforestation in general

Forests are an essential element in every ecosystem, covering 31\% of the total Earth's land area~\cite{FAOGlobal2020} and hosting most of the planet's biodiversity~\cite{VieWildlife2009}. Many species, including 80\% of amphibians, 75\% of birds, and 68\% of mammals, inhabit and depend on forested areas. 
Forests also act as a carbon sequester~\cite{HOUGHTON2023273} and as climate regulators,  influencing temperature through evapotranspiration. Furthermore, thanks to their physical structure and chemistry, forests  help mitigate extreme weather events by stabilizing rainfall patterns and limiting runoff and soil erosion~\cite{SeymourNot2022,FAOState2024}. 
These ecosystem services underscore the critical role of forests in promoting sustainable development and conserving the natural ecosystem.
However, the past few decades have witnessed a significant increase in deforestation, causing critical challenges across many biomes and exacerbating the climate change worldwide.~\cite{HOUGHTON2023273,packiam2015deforestation, PinMeoSarSal23, IPCC_2022}.  %our former work
%%Africa Suffers, NDC, EbA
Deforestation exposes soil directly to solar radiation, making it more vulnerable to weather extremes and depleting its texture, nutrients and microbial content. If these forcing factors are not addressed, they could precipitate further degradation, including desertification~\cite{ Chakravarty12, DeforCausesNASA, Geist2001WhatDT}.

Among continents, Africa had the largest annual rate of net forest loss from 2010 to 2020 (3.9 million ha lost), followed by South America with 2.6 million ha. Africa  is the continent least equipped to cope with the negative impacts of climate change. Heatwaves, heavy rains, floods, tropical cyclones, and prolonged droughts are having devastating impacts on communities and economies ~\cite{Zhao_2021, WMOAfrica2023}. As of early 2020, over a third (36\%) of all adaptation actions identified in the NDCs (Nationally Determined Contributions) of 52 African countries ~\cite{What_NDC2024} were Ecosystem-based Adaptations (EbA) ~\cite{EbA_Nalau_2018, EBA2023, EBA50ex_2024}. More than the 80\% of these actions fall within the agriculture, land use/forestry, environment and water sectors~\cite{IPCC_AR6_cap9}.
Furthermore, while the rate of net forest loss has decreased in some countries, in Africa it has continued to increase over the three decades since 1990~\cite{FAOGlobal2020}. 
Deforestation in Africa, especially in West Africa, has been frequently attributed to the agricultural activity of local farmers  ~\cite{Kummer1994, Chakravarty12, DeforCausesNASA, Geist2001WhatDT}. 
% Agriculture impacts, lash-and-burn, Deforestation
Given the low industrialization of Western African countries (e.g. Ivory Coast and Ghana), the agriculture sector is one of the major economic sectors. For many local communities farming is the only way to earn a livelihood, and then, expanding agricultural fields has become the main strategy to improve their well-being, even at the cost of deforesting many hectares~\cite{WWFForest2023}.
Movements of landless people into forests in search of farmland, along with the deplorable slash-and-burn practice — an ancient but environmentally unsustainable method for clearing forest areas for cultivation — represent some of the most significant anthropogenic causes of deforestation in tropical regions and have become a source of worldwide concern. ~\cite{packiam2015deforestation, BarimaCocoa2016,  DeforCausesNASA}.
The lack of management plans\footnote{In Africa less than 25\% of forests have management plans~\cite{FAOGlobal2020}.} and cooperation between farmers and stakeholders has posed a complex environmental, economic, and social issue which has hindered the pursuit of sustainable development~\cite{FAOGlobal2020, PinMeoSarSal23, Chakravarty12}. 

%%% CI & cocoa & needs for monitoring especially for compliances to regulations
Among the Western African countries, we decided to focus on Ivory Coast ~\cite{PinMeoSarSal23,FeMeoPinSarSal23}  %our former works on Ivory Coast
which is the largest producer of cocoa beans in the world, accounting for 43\% of the global production~\cite{kakaofacts}.
Here, cocoa production has been the main cause of the national forest loss~\cite{RufClimate2015,BarimaCocoa2016,YaoSadaiouSabasCocoa2020}.
In some areas, like in the Haut-Sassandra, deforestation has been extremely intensive with a decrease in forested area from 93\% in 1997 to 28\% in 2015~\cite{BarimaCocoa2016}. 
On average, between 1985 and 2018 more than 40\% of forests have been lost in the areas of major cocoa production~\cite{YaoSadaiouSabasCocoa2020}.
However, most of the cocoa is exported, and the European Union (EU) is one of the main per capita importers~\cite{WWFForest2023}.
Only in the last years, the EU has tried to limit the importation of products linked to deforestation activities by the introduction of deforestation-free and forest-degradation-free supply chain regulations, like the EUDR 1115/2023 ~\cite{FAOState2024, EUDR1115-2023lex, EUDR1115-2023explained}. 
However, countries like Ivory Coast may lack the monetary resources, technical solutions, and data required for compliance.
Thus, it is pivotal to develop low-cost and open-access tools to monitor forests and support management policies in developing countries.
With this aim, we have developed deep learning models to classify forest and non-forest areas using open-access satellite data. In particular, our contributions are: \textbf{a)} being the first study on Ivory Coast which has made a comparison of different deep learning models for forest segmentation; \textbf{b)} having defined a pipeline for using Sentinel open access data for developing forest segmentation data-driven models; \textbf{c)} having freely and openly released the Python codes for replications; \textbf{d)} having developed deep learning models potentially able to produce forest and deforestation maps at a finer temporal (sub-annual) resolution than existing solutions~\cite{HansenHighResolution2013,shimada2014new,jrc_map}

\section{Related Works}
\label{sec2}
% RELATED WORKS DISCOURSE:
% 1. THE IMPORTANCE OF SATELLITE IMAGES AND THEIR USE
% 2. IN THE PAST AND NOW WITH DEEP LEARNING
% 3. THE STUDY OF THE DEFORESTATION (ALL OVER THE WORLD)
% 4. IN IVORY COAST
% 5. OUR CONTRIBUTION

%%%%% DRAFT GABRIELE %%%%%
Over the years, satellite missions have gained increasing relevance in monitoring environmental phenomena, giving birth to the Earth Observation field.
For this reason, most of the countries of the world joined international missions to obtain and/or share the collected data.
Among the current most popular satellite constellations there are Sentinel~\cite{berger2012esa}, implemented by the European Space Agency, and the American constellation Landsat~\cite{wulder2019current}, which have been extensively used also thanks to their open access availability.
Satellites have been employed to monitor several aspects of the environment, such as water scarcity~\cite{ferraris2023machine}, floods~\cite{tanim2022flood}, wildfires~\cite{rashkovetsky2021wildfire}, and others \cite{mcallister2022multispectral}.

While, initially, all these satellite data were analyzed with manual methods~\cite{maretto2020spatio}, 
machine learning techniques, such as Random Forest, have been increasingly adopted for classification and segmentation~\cite{maxwell2018implementation}.
With the outbreaks of deep learning and computer vision models, a new plethora of instruments have come out. The convolutional-based architectures with their overwhelming performances have established themselves as the hard-to-beat competitor for the data-driven models. 

% SEGMENTATION DETAILS

Among targets of classification and segmentation, also deforestation gained an increasing relevance due to its impact on climate change.
Most of the researches in this field focused on the Amazon forest~\cite{bragagnolo2021amazon,john2022attention,maretto2020spatio}, but studies were carried out all over the world, over Africa~\cite{hamunyela2017using}, Europe~\cite{isaienkov2020deep,tian2013region} and Asia~\cite{wang,zhang2021deforestation}.
% HOW DEFORESTATION HAVE BEEN DONE? SPATIAL AND TEMPORAL APPROACH

% 

Deforestation has been studied both by training segmentation models on single images~\cite{bragagnolo2021amazon,john2022attention,das2023deforestation} and also by building models using a spatio-temporal approach that ingests multiple time steps~\cite{maretto2020spatio,wang2023siamhrnet}.
For instance, the first cited researches were based on an ad-hoc forest/non-forest dataset created for the Amazon forest with the support of an expert \cite{bragagnolo2019amazon_dataset}.
Consequently, these studies have been conducted only on visible images, which could be a limit to detect deforestation in areas with significant cloud coverage, which is particularly common during the summer.
Then, using Landsat images,  models have been trained on data with a spatial resolution of 30 meters per pixel.
Instead, in \cite{maretto2020spatio} the authors worked on a ground truth with a ground truth dataset distinguishing non-deforested from deforested areas, starting from PRODES dataset and creating models updating the deforested area each year.
In these case as well, 30m/px resolution Landsat images have been employed.

It is important noticing that often these segmentation tasks are possible after the creation of an ad-hoc classification map made by an expert, resulting in a time-consuming and expensive approach which is not feasible for developing countries.
Although regions like Amazon forest have been extensively studied and monitored with different datasets, Ivory Coast has been the subject of few researches~\cite{RufClimate2015,BarimaCocoa2016,YaoSadaiouSabasCocoa2020} and is still poorly covered by satellite images and specific classification maps, making it challenging to conduct research in this region.
This research aims to train state-of-the-art deep learning models on open-access datasets to perform a forest/non-forest classification in the Ivory Coast.

In this paper, different models are compared using different inputs, such as visible and radar acquisitions, both separately and jointly.
Since  deforestation can often be detected through visual inspection of RGB images, most of the cited researches are based on this type of data images~\cite{bragagnolo2021amazon,john2022attention,das2023deforestation,wang2023siamhrnet}.
However, to deal with the particularly high probability of cloud coverage over the forest during the year, this study also evaluates the use of radar acquisitions, which are not affected by atmospheric conditions.
For this purpose, Sentinel-1 and Sentinel-2 are employed with the global Forest/Non-Forest map (FNF)~\cite{shimada2014new} to create free models scalable worldwide. 
Finally, the model's classification map predictions are used to detect deforestation by looking at the changes in per-pixel classifications over a given period.

%%%%% DRAFT GABIRELE %%%%%

\begin{figure*}[ht]
    \centering
\includegraphics[width=0.975\textwidth]{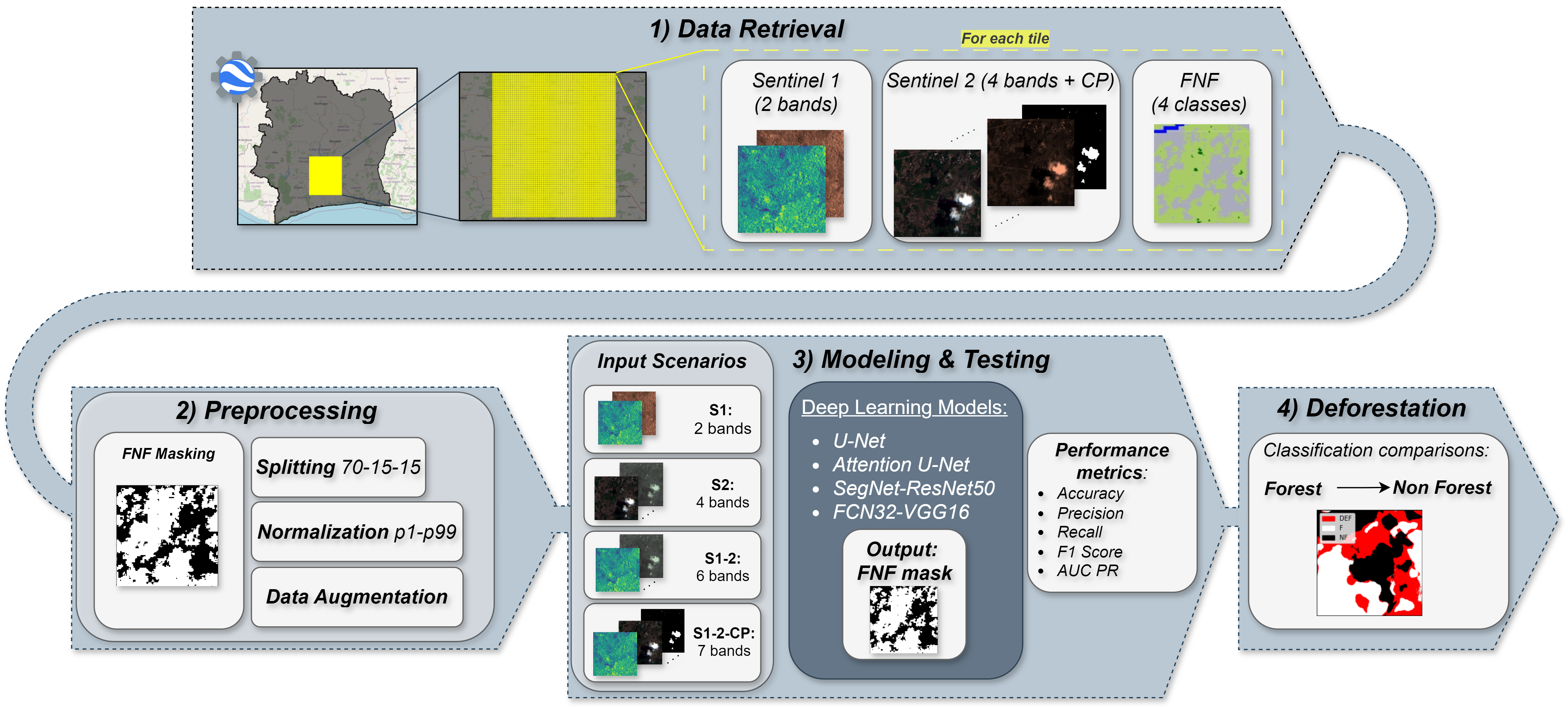}
    \caption{The pipeline of the system: 1) \textbf{data retrieval}, exploiting Google Earth Engine Python API to select and download data, 2) \textbf{preprocessing}, splitting, normalizing and augmenting data 3) \textbf{modelling \& testing} and 4) \textbf{deforestation} calculus.}
    \label{fig:pipeline}
\end{figure*}

\section{Datasets and Methods}
\label{sec3}
This work aims to provide a methodology to develop deforestation detectors over rural areas scarcely covered by remote sensing open source data, such as the Ivory Coast.
The pipeline performed in this work, and depicted in Figure \ref{fig:pipeline}, is the following:
% TROPPO GENERICO, MEGLIO UN'IMMAGINE
\begin{enumerate}
    \item \textbf{Data retrieval:} selection of the region of interest (ROI), mosaic definition, crop raw data into tiles, per-tiles download of Sentinel-1 (2 bands), Sentinel-2 (4 bands and Cloud Probability), and FNF map;
    \item \textbf{Preprocessing}: ground truth processing, splitting of the dataset into train, validation and test sets, normalization and data augmentation;
    \item \textbf{Modeling \& Testing:} models initialization, training, and testing using some performance metrics on test sets;
    \item \textbf{Deforestation:} Detect deforestation in a given period by looking at the changes in the forest classification map predictions.
\end{enumerate}
Step \textbf{1)} was performed on Google Earth Engine, while the remaining steps were performed through Python using Cineca HPC resources \footnote{https://www.hpc.cineca.it/}. 
The following subsections provide more details about these steps.

\subsection{Datasets}\label{subsec3_1}
% ASSUMIAMO CHE IL PERCHé SIA GIà STATO SPIEGATO
% Assumiamo un nome per il nostro dataset? Perché anche gli altri sono dataset, magari S2FNF?
The raw data used in this work consists of  Sentinel datasets and the Forest/Non-Forest (FNF) dataset, provided by the European Copernicus program and the Japanese ALOS program, respectively.
The Sentinel data are used as input, while the FNF classification is the ground truth.
To avoid ambiguity, in the rest of the paper, the created dataset made of instancies of Sentinel and FNF images is referred to as S2FNF. % più scorrevole (?): To avoid ambiguity, the dataset created from instances of Sentinel and FNF images will be referred to as S2FNF in the rest of the paper.

\subsubsection{Sentinel}\label{subsbusec3_1_1}
Sentinel-1 and Sentinel-2 are two Earth Observation missions of the Copernicus program, managed by the European Commission.
Sentinel-1 carries a C-band synthetic-aperture radar (SAR) instrument, while Sentinel-2 provides 13-bands multispectral images.
%, both with a maximum resolution of 10 meters per pixel (m/px).
These two missions offer complementary data:
 Sentinel-1 captures images in the non-visible electromagnetic spectrum which are not influenced by time of acquisition (day/night) and weather conditions.
 While, Sentinel-2 collects images primarily in the visible spectrum helpful for monitoring Earth's land and coastal waters.
Sentinel images were chosen as the input of the proposed models because of their sub-monthly revisit time and  spatial resolution of 10 meters per pixel (m/px) which make these data highly suitable for the segmentation task.

\subsubsection{Forest-Non-Forest map (FNF)}\label{subsbusec3_1_2}
The target for training our model is the Forest/Non-Forest map (FNF version 4)~\cite{shimada2014new}.
This map is produced by processing the Advanced Land Observing Satellite Phased Arrayed L-band Synthetic Aperture Radar (ALOS-PALSAR) data.
The FNF map has a native resolution of 25m/px and classifies pixels into four categories: dense forest, non-dense forest, non-forest and water.
To align the input and output dimensions of the models, we upsampled FNF images to match the Sentinel's resolution of 10m/px.
Furthermore, following the Food and Agriculture Organization (FAO) forest definition\footnote{According to the FAO, forests are defined as \textit{``lands of more than 0.5 hectares, with a tree canopy cover of more than 10\%, which are not primarily under agricultural or urban land use"}. For more details refer to: https://www.fao.org/4/ad665e/ad665e03.htm}, we consolidated the original 4 categories into two: ``forest" (made of dense and non-dense forest) and ``non-forest" (made of non-forest and water).
In this classification, ``forest" is the positive label.
Other maps classifying forests, such as ESA World Cover ~\cite{esa_worldcover_2021} and JRC Global Forest map ~\cite{jrc_map}), have been released.
However, the FNF map offers a broader temporal coverage (since 2007 up to the present) and  is specifically tailored for forestry classifications.

\subsubsection{S2FNF dataset}
\label{sec3_1_3}
\begin{figure}[ht]
    \centering
\includegraphics[width=0.45\textwidth]{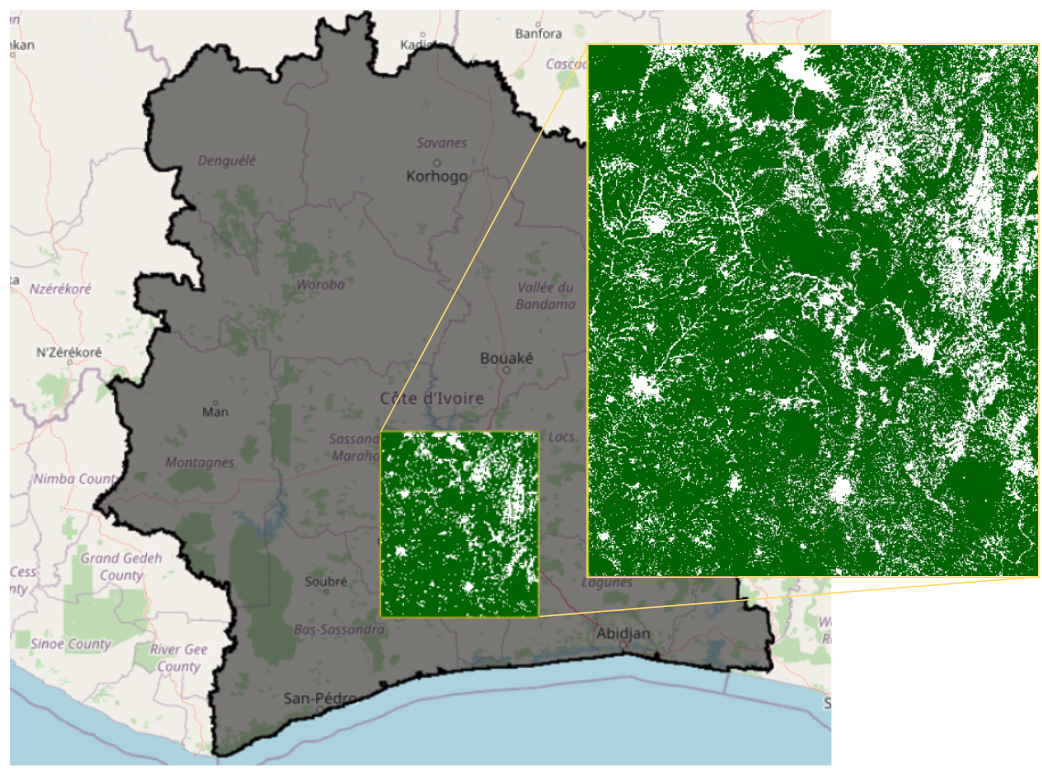}
    \caption{Forest (including dense and non-dense forest) of the ROI according to the FNF.}
    \label{fig:roi}
\end{figure}

To reduce computational load, we focused on a representative region in the central part of the country, retrieving data from 2019. The region of interest (ROI) %acronym has been expanded
is defined by longitudes ranging from -6.0969 to -4.7731  and latitudes from 5.5697 to  7.1474 (see Figure~\ref{fig:roi}).

\begin{figure}[ht]
    \centering
\includegraphics[width=0.47\textwidth]{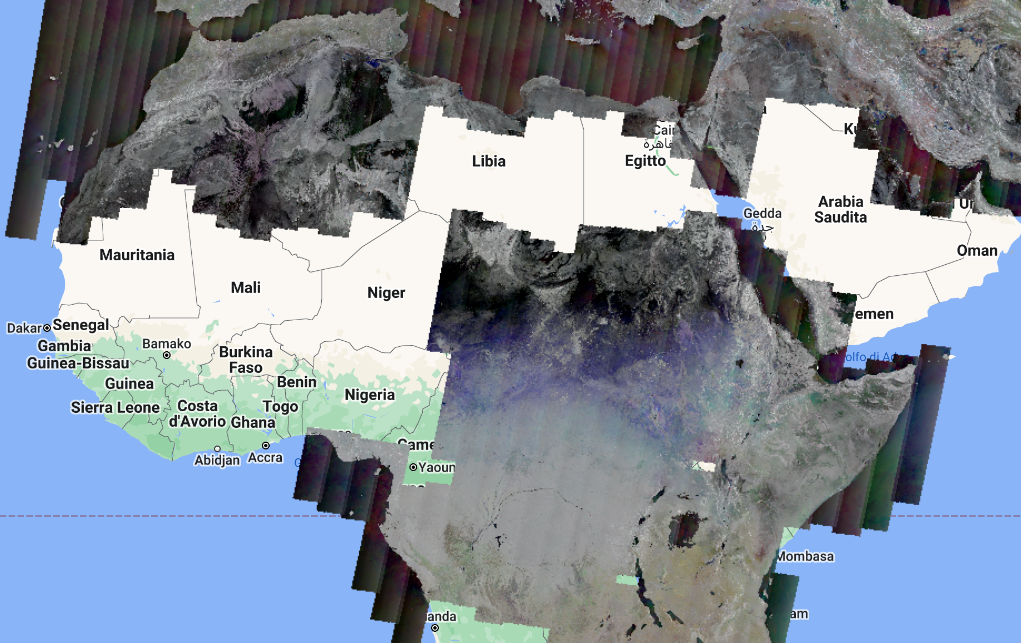}
    \caption{Sentinel-1 descending orbit images during 2019 displayed on Google Earth Engine. No image is available over Ivory Coast.}
    \label{fig:sent1_coverage}
\end{figure}

Sentinel's mosaics have been composed by images for the period from May to October 2019.
This period was chosen because the FNF map for this region is generated yearly by %using %  
ALOS-PALSAR data from the same months. 
Thus, in this way, the input data should be the most representative of the target ground-truth map.  %forse meglio (?): By aligning the input data with this timeframe, we ensure that data are the most representative of the target ground-truth map
Over the ROI only VV (single co-polarization, vertical transmit/vertical receive) and VH (dual-band cross-polarization, vertical transmit/horizontal receive) bands of Sentinel-1 in ascending orbit were % invece che "have been" 
available. 
%Given the context where the data from 2019 were used for research conducted in the present (2024), the simple past tense ("were available") is appropriate. This is because you are referring to the specific availability of data from a past point in time (2019) that you accessed during your research.
As can be seen in Figure~\ref{fig:sent1_coverage} there is no descending orbit data for Sentinel-1 in 2019, and this is also true for other years.
For Sentinel-2, four bands have been retrieved: B2, B3, B4 and B8, respectively, blue, green, red and NIR (i.e. visible and near-infrared, also known as RGBN).
These bands have a resolution of 10m/px which is the finest resolution available among the %other (other cancellato) 
Sentinel-2 bands\footnote{Details about Sentinel-2 bands can be found here: \url{https://sentiwiki.copernicus.eu/web/s2-mission}.}).
Given that summer is the season presenting extensive cloud coverage in Ivory Coast, Sentinel-2 images has been filtered to include only those with at most 20\% of cloud coverage.
Even though this filtering prevents the use of almost completely cloud-covered images  %forse non è  chiarissimo "almost completely cloud-coverage images"; si intended qualcosa come "Although this filtering excludes nearly all the cloud-covered images"  ?
, many of the remaining Sentinel-2 images still suffer from cloud occlusion which makes the segmentation task particularly challenging using only Sentinel-2 optical data. For this reason, we decided to use also the Cloud Probability Map layer provided within the Sentinel-2 product, which represents the probability of each pixel being covered by clouds. 

The ROI has been mosaiced % divided? 
into 4000 squared tiles, each covering an area approximately of 6,55 km$^2$. 
%side 2560 meters long.
The S2FNF has been composed then by 4000 instances, each made of Sentinel features and the respective FNF target mask.
The S2FNF has been split into 70\% training, 15\% validation and 15\% test sets.
Finally, it is worth noting that the Ivory Coast is mostly covered by forests and, consequently, the dataset classification labels are significantly unbalanced towards this class.
Specifically, around 75\% of the labels are forest areas, while only 25\% are non-forest.
For this reason, during training, the loss is weighted in an inversely proportional way, weighting 0.7 the misclassification of a non-forest pixel and 0.3 the forest misclassification.
% (? better ?): For this reason,  during training, the loss is weighted inversely to address this imbalance: misclassification of non-forest pixels is weighted at 0.7, while forest misclassification is weighted at 0.3.

\subsection{Preprocessing}\label{sec3_2}
% The input of S2FNF, composed by the Setinel's images, has been preprocessed to deal with outlier values.
Sentinel images have been normalized following the Equation~\ref{eq:p_norm} which is the classical min-max normalization but using instead the 1st and 99th percentile. This improves the representation of features in the normalized space in the presence of outliers,  easing in this way  the information extraction and improving the learning process of the models.

\begin{equation}
    X = (p_{99} - X) / (p_{99} - p_{1}).
\label{eq:p_norm}
\end{equation}

% ANY EVIDENCE OF IT?
Data augmentation is applied at training time\footnote{Augmentation is done using the ImageDataGenerator class of Keras.}.
This process does not create a specific percentage of new samples, but it slightly modifies all the samples during training.
Specifically, it applies random horizontal/vertical flips, shifts (from 0 to 10\% of the width/height length), and rotations (from 0 to 180 degrees). 

\subsection{Models}\label{sec3_3}

The experiments  are carried out comparing four segmentation models: FCN32~\cite{long_fully_2015} with a VGG16~\cite{simonyan2015vgg} backbone, SegNet~\cite{badrinarayanan2016segnet} with a ResNet50~\cite{heresnet2015} backbone, UNet~\cite{ronneberger2015u}, and the Attention UNet proposed by~\cite{oktay2018attunet}. 
The term backbone is generally used when multiple architectures are employed for building a single final model.
As the name suggests, the backbone is the architecture which characterizes the inner structure of a model (e.g. VGG) and works as a feature extractor, while other architectural structures could be used to define the shape of the final model (e.g. FCN).
To develop specific forest classifiers, all the architectures have been trained from scratch without using transfer learning.
While FCN, SegNet and UNet are state-of-the-art and very widespread models, Attention-UNet has been less adopted except in the medical field.
In more detail, Attention-UNet is a classical UNet with the attention gates along with the skip connection of the UNet, these gates should help the network to focus only on salient features and suppress irrelevant regions of the feature maps. 
In \cite{john2022attention} authors highlighted the performance of this architecture for classifying forests in South America, and for this reason, we have decided to add it to our comparison.

% LET'S SPEAK ALSO ABOUT TRIANING HERE
\section{Results}
\label{sec4}

\subsection{Forest Segmentation}
Experiments have been carried out considering four input combination scenarios: \textbf{S1}) Sentinel-1, \textbf{S2}) Sentinel-2, \textbf{S1-2}) Sentinel-1 and Sentinel-2 combined, \textbf{S1-2-CP}) Sentinel-1, Sentinel-2, and Cloud Probability Map combined.
All the models have been trained using the Adam optimizer with a binary cross-entropy loss for $50$ epochs with a learning rate of $0.0001$ and a batch size of $32$.
Two test sets have been defined to evaluate the generalization ability of the models, especially concerning cloud coverage. The first test set was defined in 2019, the same year as the training, in which the cloud coverage over the ROI was about $10.9$\%; the second test set was defined in 2020, in which the cloud cover was about $6.8$\% over the ROI.
If a model can generalize well and is resilient to adverse weather conditions, its performances should be maintained over the test set defined in the two consecutive years.
Accuracy, precision, recall, f1-score, and AUC PR (AUC of PR curves) have been computed to evaluate the performance of the models. 

% NO ROC perchè imbalanced~\cite{}(Takaya Saito)

% It is important to note that, in contrast with the training in which the misclassification cost is weighted as described in section \ref{sec3_1_3}, the evaluation metrics of the test set is calculated without applying any weights to focus on a specific label.

\begin{table*}[ht!]
    \centering
    \resizebox{\textwidth}{!}{%
    \begin{tabular}{lcccccccccc}
        \toprule
        \multirow{2}{*}{Classifier} & \multicolumn{5}{c}{Test 2019} & \multicolumn{5}{c}{Test 2020} \\ 
        \cmidrule(lr){2-6} \cmidrule(lr){7-11}
        & Accuracy & Precision & Recall & F1\_Score & AUC PR & Accuracy & Precision & Recall & F1\_Score & AUC PR \\
        \midrule
        U-Net           & \textbf{0.8821} & \textbf{0.9073} & 0.9067 & 0.9026 & \textbf{0.8828} & 0.8811 & \textbf{0.9013} & 0.9072 & 0.9002 & \textbf{0.8781}  \\
        \hline
        Attention U-Net & 0.8777 & 0.8984 & 0.9095 & 0.8998 & 0.8768 & 0.8770 & 0.8918 & 0.9129 & 0.8980 & 0.8713 \\
        \hline
        SegNet-ResNet50 & 0.8817 & 0.9000 & 0.9148 & \textbf{0.9038} & 0.8789 & \textbf{0.8830} & 0.8938 & \textbf{0.9198} & \textbf{0.9032} & 0.8738 \\
        \hline
        FCN32-VGG16     & 0.8592 & 0.8679 & \textbf{0.9172} & 0.8869 & 0.8494 & 0.8599 & 0.8609 & 0.9205 & 0.8846 & 0.8421 \\
        \bottomrule
    \end{tabular}%
    }
    \caption{\textbf{S1} scenario, performance metrics computed on the test sets in 2019 and 2020.}
    \label{tab:perf_s1}
\end{table*}

\begin{table*}[ht!]
    \centering
    \resizebox{\textwidth}{!}{%
    \begin{tabular}{lcccccccccc}
        \toprule
        \multirow{2}{*}{Classifier} & \multicolumn{5}{c}{Test 2019} & \multicolumn{5}{c}{Test 2020} \\ 
        \cmidrule(lr){2-6} \cmidrule(lr){7-11}
        & Accuracy & Precision & Recall & F1\_Score & AUC PR & Accuracy & Precision & Recall & F1\_Score & AUC PR \\
        \midrule
        U-Net           & \textbf{0.8613} & 0.8945 & 0.8923 & \textbf{0.8878} & 0.8696 & \textbf{0.8577} & 0.8675 & \textbf{0.9161} & \textbf{0.8852} & 0.8503 \\
        \hline
        Attention U-Net & 0.8623 & 0.8902 & \textbf{0.8969} & 0.8870 & 0.8658 & 0.8376 & 0.8411 & 0.9150 & 0.8697 & 0.8263 \\
        \hline
        SegNet-ResNet50 & 0.8581 & \textbf{0.9044} & 0.8700 & 0.8781 & \textbf{0.8707} & 0.8343 & \textbf{0.8902} & 0.8423 & 0.8570 & \textbf{0.8552} \\
        \hline
        FCN32-VGG16     & 0.8377 & 0.8829 & 0.8592 & 0.8605 & 0.8499 & 0.8362 & 0.8444 & 0.8986 & 0.8636 & 0.8225 \\
        \bottomrule
    \end{tabular}%
    }
    \caption{\textbf{S2} scenario, performance metrics computed on the test sets in 2019 and 2020.}
    \label{tab:perf_s2}
\end{table*}

\begin{table*}[ht!]
    \centering
    \resizebox{\textwidth}{!}{%
    \begin{tabular}{lcccccccccc}
        \toprule
        \multirow{2}{*}{Classifier} & \multicolumn{5}{c}{Test 2019} & \multicolumn{5}{c}{Test 2020} \\ 
        \cmidrule(lr){2-6} \cmidrule(lr){7-11}
        & Accuracy & Precision & Recall & F1\_Score & AUC PR & Accuracy & Precision & Recall & F1\_Score & AUC PR \\
        \midrule
        U-Net           & 0.8864 & \textbf{0.9133} & 0.9036 & 0.9045 & \textbf{0.8882} & \textbf{0.8801} & \textbf{0.8933} & \textbf{0.9124} & \textbf{0.8971} & 0.8685 \\
        \hline
        Attention U-Net & 0.8760 & 0.9021 & 0.9025 & 0.8983 & 0.8793 & 0.8599 & 0.8554 & 0.9357 & 0.8878 & 0.8406 \\
        \hline
        SegNet-ResNet50 & \textbf{0.8871} & 0.9090 & \textbf{0.9117} & \textbf{0.9070} & 0.8859 & 0.8727 & 0.8924 & 0.9020 & 0.8923 & \textbf{0.8675} \\
        \hline
        FCN32-VGG16     & 0.8626 & 0.8917 & 0.8889 & 0.8852 & 0.8642 & 0.8619 & 0.8831 & 0.8902 & 0.8804 & 0.8538 \\
        \bottomrule
    \end{tabular}%
    }
    \caption{\textbf{S1-2} scenario, performance metrics computed on the test sets in 2019 and 2020.}
    \label{tab:perf_s1_2}
\end{table*}

\begin{table*}[ht!]
    \centering
    \resizebox{\textwidth}{!}{%
    \begin{tabular}{lcccccccccc}
        \toprule
        \multirow{2}{*}{Classifier} & \multicolumn{5}{c}{Test 2019} & \multicolumn{5}{c}{Test 2020} \\ 
        \cmidrule(lr){2-6} \cmidrule(lr){7-11}
        & Accuracy & Precision & Recall & F1\_Score & AUC PR & Accuracy & Precision & Recall & F1\_Score & AUC PR \\
        \midrule
        U-Net           & \textbf{0.8877} & 0.8993 & \textbf{0.9258} & \textbf{0.9095} & 0.8808 & \textbf{0.8750} & 0.8823 & \textbf{0.9213} & \textbf{0.8964} & 0.8612 \\
        \hline
        Attention U-Net & 0.8751 & 0.8895 & 0.9188 & 0.9005 & 0.8719 & 0.8717 & \textbf{0.8988} & 0.8911 & 0.8895 & 0.8703 \\
        \hline
        SegNet-ResNet50 & 0.8854 & \textbf{0.9133} & 0.9021 & 0.9036 & \textbf{0.8869} & 0.8528 & 0.9148 & 0.8382 & 0.8642 & \textbf{0.8706} \\
        \hline
        FCN32-VGG16     & 0.8640 & 0.8839 & 0.8996 & 0.8873 & 0.8597 & 0.8616 & 0.8687 & 0.9053 & 0.8806 & 0.8442 \\
        \bottomrule
    \end{tabular}%
    }
    \caption{\textbf{S1-2-CP} scenario, performance metrics computed on the test sets in 2019 and 2020.}
    \label{tab:perf_s1_2_cp}
\end{table*}

\subsubsection{Sentinel-1 (S-1)}\label{sec4_1}
\begin{figure*}[ht!]
    \centering
\includegraphics[width=0.95\textwidth]{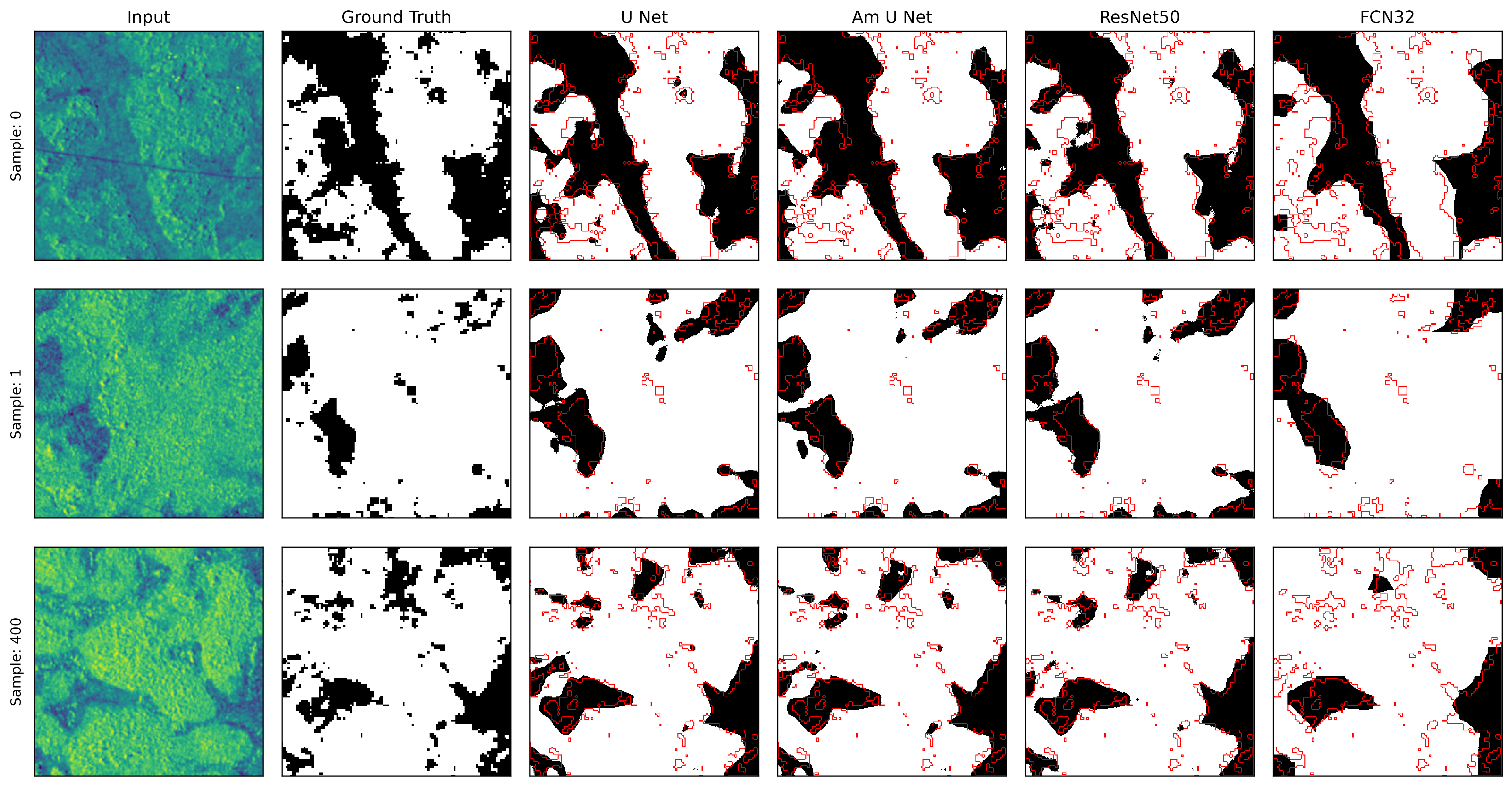}
    \caption{Models' prediction samples in the \textbf{S1} scenario in 2019 test. Each row shows the input and the ground truth of a specific instance. Then, from the third column (from left to right) models' predictions are displayed. The red overlay shows the ground truth borders.}
    \label{fig:sent1_pred_mask}
\end{figure*}

In this scenario, models have been fed using only  Sentinel-1 bands, thus the predictions should not be affected by cloud coverage.
Among all the models, U-Net and SegNet-ResNet50 have performed the best % exhibited the best performances, with quite similar results
with quite similar performances (Table~\ref{tab:perf_s1}). 
Figure~\ref{fig:sent1_pred_mask} represents the classification prediction of the different models in the \textbf{S1} scenario. 
All the metrics  for the two test sets are almost the same 
in consecutive years % meglio: across consecutive years
, suggesting that these models can generalize well and are not affected by cloud coverage - as expected and explained %meglio : and discussed
in Section~\ref{sec3}. 
The trade-off between precision and recall appears to be well balanced with an F1 score of around 0.9 for the best-performing models.
In the \textbf{S1} scenario, the recalls are generally and slightly higher than the precision values; this could be attributed to the dataset imbalance whose positive class (forest) is more frequent % meglio: where positive class (forest) is more frequent.
All the above considerations are equally valid either for the 2019 and 2020 test sets, demonstrating the resilience %forse meglio: robustness
 of using Sentinel-1 data in different circumstances. %forse meglio: conditions; scenarios. 

%SP corrections STOPs here

\begin{figure*}[ht!]
    \centering
\includegraphics[width=0.95\textwidth]{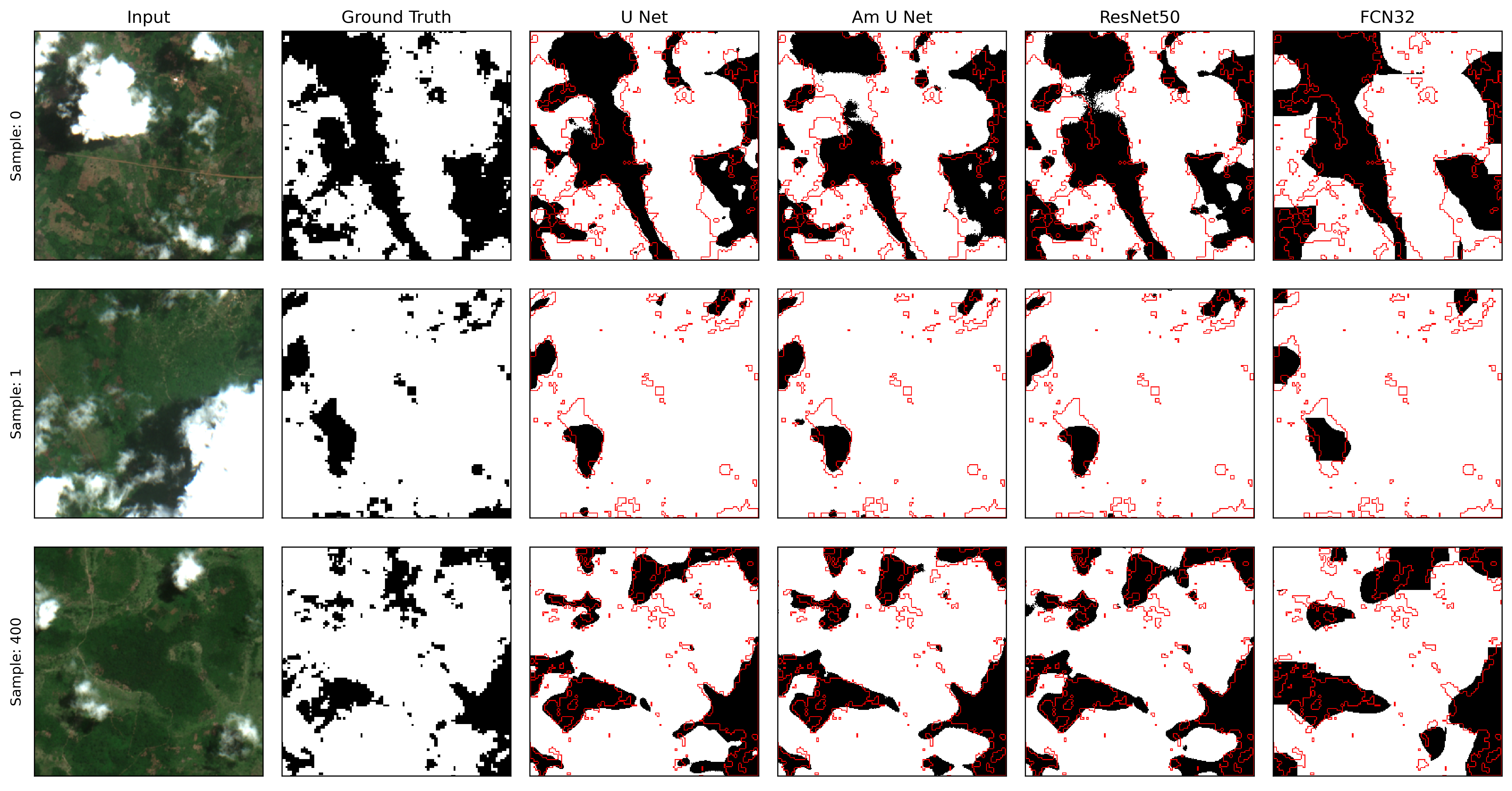}
    \caption{Models' prediction samples in the \textbf{S2} scenario in 2019 test. Each row shows the input and the ground truth of a specific instance. Then, from the third column (from left to right) models' predictions are displayed. The red overlay shows the ground truth borders.}
    \label{fig:sent2_pred_mask}
\end{figure*}

\subsubsection{Sentinel-2 (S2)}\label{sec4_2}

In this scenario, models have been trained using Sentinel-2 bands as input.
The best-performing models (Table~\ref{tab:perf_s2}) remain U-Net and SegNet-ResNet50 respectively.
Comparing precision and recall in the \textbf{S2} scenario in the 2019 test, the precisions are slightly higher than recalls (more false negative than false positive, i.e. more false non-forest than false forest) for three models (U-Net, SegNet-ResNet50, FCN-VGG16) over four.
However, in the 2020 test, among these three models, only SegNet-ResNet shows a precision higher than the recall.
Considering that 2020 has a lower cloud coverage, and the detrimental effect of clouds on the information content of optical data, this might mean that U-Net and FCN-VGG16 in \textbf{S2} scenario are prone to classify clouds as non-forest erroneously.
Relating to the previous \textbf{S1} scenario, performance metrics are worse both for the tests in 2019 and 2020.
%This could also be seen from the prediction in Figure~\ref{fig:sent2_pred_mask}.
The difference in the AUC PR in the 2020 test between \textbf{S1} and \textbf{S2} highlights and states the limitation of developing models based only on optical Sentinel-2 data on a cloudy region.
In this context, SAR data could be considered as input features to complete the noisy (partially cloud-occluded), or even missing (totally cloud-occluded), optical land satellite images.

\subsubsection{Sentinel-1 \& Sentinel-2 (S1-2)}\label{sec4_3}

Given the limitation of the \textbf{S2} scenario in cloudy context, we have proposed a new scenario combining Sentinel-1 and Sentinel-2 data with the idea of exploiting the most useful information from both visible and non-visible sources.
Indeed, as already stated, in case of highly cloud occluded regions, SAR data should be considered the main feature for the classification task.
However, optical data are for sure the faster and easiest means by which performing forest and non forest segmentation in a cloud-free context. 
This scenario is named \textbf{S1-2} and provides the concatenation of Sentinel-1 and Sentinel-2 bands (i.e., 6 bands, namely VV, VH, B2, B3, B4, B8) as the input of the models.
On the 2019 test, models in the \textbf{S1-2} scenario seems to perform slightly better than \textbf{S1} in terms of accuracy and AUC PR (Table~\ref{tab:perf_s1_2}). 
Accuracies and recalls are still well-balanced, with a F1 score around 0.9 for all the models.
However, on 2020 test, models perform not better, and in some case worse, than in \textbf{S1}.
This suggest that in \textbf{S1-2} scenario models generalize less than in \textbf{S1}, probably beacuse of the inability of
models to understand on what sources to focus on depending on cloud coverage. 
This is demonstrated also by the higher AUC PR in the \textbf{S1} scenario (Table~\ref{tab:perf_s1}). 

\subsubsection{Sentinel-1 \& Sentinel-2 \& Cloud Probability (S1-2-CP)}\label{sec4_4}

To facilitate the learning of how to use Sentinel-1 and Sentinel-2, we have proposed a new scenario (\textbf{S1-2-CP}) in which the Cloud Probability Map is provided as an additional input to the \textbf{S1-2} features.
Models' performances (Table~\ref{tab:perf_s1_2_cp}) on the 2019 test do not reveal any significative improvement than the \textbf{S1-2} scenario, and for some models and metrics even worse.
This suggest that \textbf{S1} scenario provides sufficient information for the classification task and the proposed integration of optical data sem to be unnecessary - even misleading in some occasions.
The high cloud coverage represents for sure one of the major limiting factor for the usefulness of Sentinel-2 data. 
More complex models could be developed and tested to ease the automatic fusion of SAR and optical data, as discussed in Section~\ref{sec5}.

\subsection{Deforestation detection}

Concerning the forest and non-forest segmentation task, U-Net has revealed the best-performing model. 

\begin{figure*}[h!]
    \centering
\includegraphics[width=0.7\textwidth]{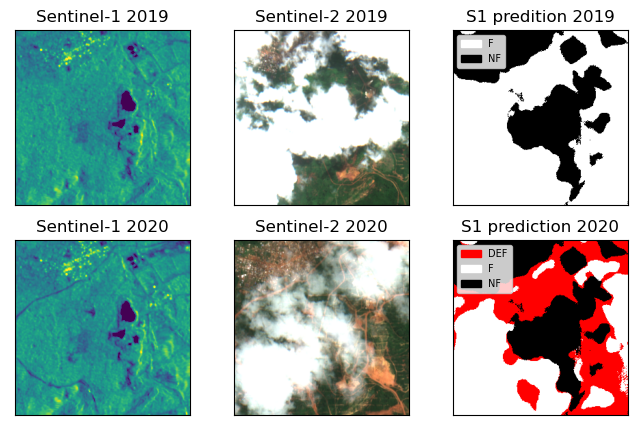}
    \caption{(In the first row (from left to right): Sentinel-1 VV bands for 2019, Sentinel-2 data for 2019, U-Net segmentation map prediction for 2019 in the \textbf{S1} scenario.
    In the second row (from left to right): Sentinel-1 VV bands for 2020, Sentinel-2 data for 2020, U-Net segmentation map prediction for 2020 in the \textbf{S1} scenario highlighting in red non-forest pixel which were forest in 2019 (i.e. deforested pixel).}
    \label{fig:sent1_def_prediction}
\end{figure*}

For this reason, we have decided to select U-Net in the \textbf{S1} scenario as the reference for deforestation mapping.
Our strategy to detect deforestation has been to look at changes in per-pixel classification in two time-subsequent segmentation maps: when a forest pixel becomes a non-forest pixel deforestation has occurred. 
Figure~\ref{fig:sent1_def_prediction} shows the predicted deforestation between 2019 and 2020 for a sample tile.
The usefulness of Sentinel-1 SAR data is definitively demonstrated here.
Indeed, both Sentinel-2 images in 2019 and 2020 are partially occluded by clouds, and understanding deforestation patterns is rather complicated.
Instead, looking at radar data, it is clear the urban expansion and the consequent deforestation.
It is relevant to stress that deforestation maps like in Figure~\ref{fig:sent1_def_prediction} could be created for any tile and periods with any length, where the lower bound length is defined by the Sentinel-1 revisit time which is approximately two weeks. 
Looking for deforestation all over the ROI, we have estimated 462.52 $km^2$ of deforested area between 2019 and 2020, which means deforestation of roughly 2\% concerning 2019 forests. 

\section{Discussion}
\label{sec5}

Focusing on the forest segmentation task, one of the major problems has been the difficulty of correctly classifying isolated non-forest pixels, as it is visible from Figure~\ref{fig:sent1_pred_mask} and Figure~\ref{fig:sent2_pred_mask}.
This is a rather difficult goal, even in ideal conditions of high-quality data and well-performing models.

To the best of our knowledge, no previous works have been done applying deep learning models over the Ivory Coast for forest/non-forest segmentation, especially using FNF.
However, other works on the same task presented models reaching about 95\% of accuracy and F1-score ~\cite{bragagnolo2021amazon,john2022attention,das2023deforestation}.
Instead, although the models presented in this paper reached about 90\% of accuracy, it is important noticing that our ground truth could be less precise.
Indeed, in Ivory Coast, the conditions are far from ideal, especially because of the data availability and quality both for the target FNF map and for the Sentinel images. 
Indeed, FNF's documentation declares an accuracy of 91\% for its forest classification over Africa.
This means that the target used for training deep learning models might be erroneous in some circumstances, hardening the model training.
Furthermore, FNF has an original spatial resolution of 25m/px, which is lower than the Sentinel images one.
Consequently, models could have learned well how to classify forest and non-forest classes, but FNF could act as a bottleneck for models performances on this task.
In other words, the metrics of Table \ref{tab:perf_s1}, \ref{tab:perf_s2}, \ref{tab:perf_s1_2}, and \ref{tab:perf_s1_2_cp} could be considered as lower bounds, being potentially higher in the case in which FNF map would increase its resolution up to 10m/px.

The major limitation of Sentinel-1 data is the lack of descending orbit images. Even if our \textbf{S1} scenario has produced remarkable results, the availability of this additional Sentinel-1 data could enhance a lot the ability of models to classify forest and non-forest highlighting additional forest borders - which in turn would bring more accurate deforestation maps.

Notwithstanding the limitations, we have provided an open-access framework to produce deforestation maps in Ivory Coast using only open data.
As already pointed out, within our framework, deforestation maps could be created at a sub-annual rate, which is a temporal frequency higher than any other existing tool. 
Furthermore, by using Sentinel-1 data we have developed cloud-resilient remote sensing-based models, which is essential for countries in which clouds represent a limiting factor for optical satellite data.

% FUTURE WORKS
In this case study the combination of both Sentinel-1 and Sentinel-2 has been of scarce utility. However, more complex models could be developed aiming at a better automatic information fusion in the \textbf{S1-2} scenario. In detail, the so called ``neuro-symbolic" or ``physically-informed" approach~\cite{de2019deep} could be used to force the network to learn autonomously where clouds are present and then, according to cloud occlusion intensity, what source between Sentinel-1 and 2 to use more.
This improvement is left for future work.

To improve the proposed framework, a spatio-temporal approach could be tested.
Instead of training models for forest segmentation, and subsequently, looking for per-pixel label changes in a deterministic way, it is possible to develop models that directly predict deforestation patterns.
However, this approach would require feeding as input two satellite images, one at the beginning of the period to monitor, and one at the end.
Furthermore, a true deforestation map should be available, i.e. a ground truth. 
Most of the studies retrieve this ground truth either from pre-existing datasets or by expert manual labelling, which is time and cost-intensive~\cite{maretto2020spatio,wang2023siamhrnet}.
West Africa, and in particular Ivory Coast, lacks such datasets and given the scarce resources we have preferred to produce a simpler and cheaper, but re-usable and still accurate, instrument by adopting the FNF and Sentinel data.

Finally, other analysis will be carried out over the rest of the Ivory Coast and other similar countries to validate further our results.

\section{Conclusions}
\label{sec6}

Despite the availability of remote sensing open data, this analysis highlights the difficulty of conducting a study on developing countries with scarce economic and technical resources.
It is pivotal to provide cheap, or even free, instruments to these countries to comply with standards and international regulations on the environmental impact of goods.
Our study has tried to furnish a free deep-learning model to detect deforestation patterns in the Ivory Coast, demonstrating the potential and the usefulness of radar acquisitions over forested areas.
Even if some limitation exists and multiple improvements are applicable, our solution could be considered a first approach to support forest sub-annual monitoring and national forest management policies.


%\appendix
%\section{Results 2019}
%\label{app1}


\bibliographystyle{elsarticle-num}
\bibliography{bibliography}

\end{document}